\documentclass[11pt,a4paper]{article}
\usepackage[hyperref]{acl2021}
\usepackage{times}
\usepackage{latexsym}
\usepackage{kky}

\usepackage{graphicx}
\usepackage{amsmath}
\usepackage{booktabs}
\usepackage{amsfonts,amssymb}
\usepackage{amssymb}
\usepackage{url}
\usepackage{enumitem}
\usepackage{multirow}

\usepackage{microtype}

\aclfinalcopy % Uncomment this line for the final submission
\title{ROPE: Reading Order Equivariant Positional Encoding\\ for Graph-based Document Information Extraction}

\author{Chen-Yu Lee\textsuperscript{$\dagger$}, Chun-Liang Li\textsuperscript{$\dagger$}, Chu Wang\textsuperscript{$\mathsection$}\thanks{\indent{Work done while an intern at Google Research.}}, Renshen Wang\textsuperscript{$\ddagger$}, \\ \textbf{Yasuhisa Fujii\textsuperscript{$\ddagger$},  Siyang Qin\textsuperscript{$\ddagger$}, Ashok Popat\textsuperscript{$\ddagger$}, Tomas Pfister\textsuperscript{$\dagger$}} \\
  \textsuperscript{$\dagger$}Google Cloud AI, \textsuperscript{$\mathsection$}McGill University, \textsuperscript{$\ddagger$} Google Research \\
  \texttt{\textsuperscript{$\dagger$,$\ddagger$}\{\footnotesize{chenyulee, chunliang, rewang, yasuhisaf, qinb, popat, tpfister}\}@google.com} \\ \texttt{\textsuperscript{$\mathsection$}\footnotesize{chu.wang@mail.mcgill.ca}} \\}

\date{}

\begin{document}
\maketitle
\begin{abstract}
Natural reading orders of words are crucial for information extraction from form-like documents. Despite recent advances in Graph Convolutional Networks (GCNs) on modeling spatial layout patterns of documents, they have limited ability to capture reading orders of given word-level node representations in a graph. We propose Reading Order Equivariant Positional Encoding (ROPE), a new positional encoding technique designed to apprehend the sequential presentation of words in documents. ROPE generates unique reading order codes for neighboring words relative to the target word given a word-level graph connectivity. We study two fundamental document entity extraction tasks including word labeling and word grouping on the public FUNSD dataset and a large-scale payment dataset. We show that ROPE consistently improves existing GCNs with a margin up to 8.4\% F1-score.
\end{abstract}

\section{Introduction}
\begin{figure}[!t]
\centering
\includegraphics[width=0.47\textwidth]{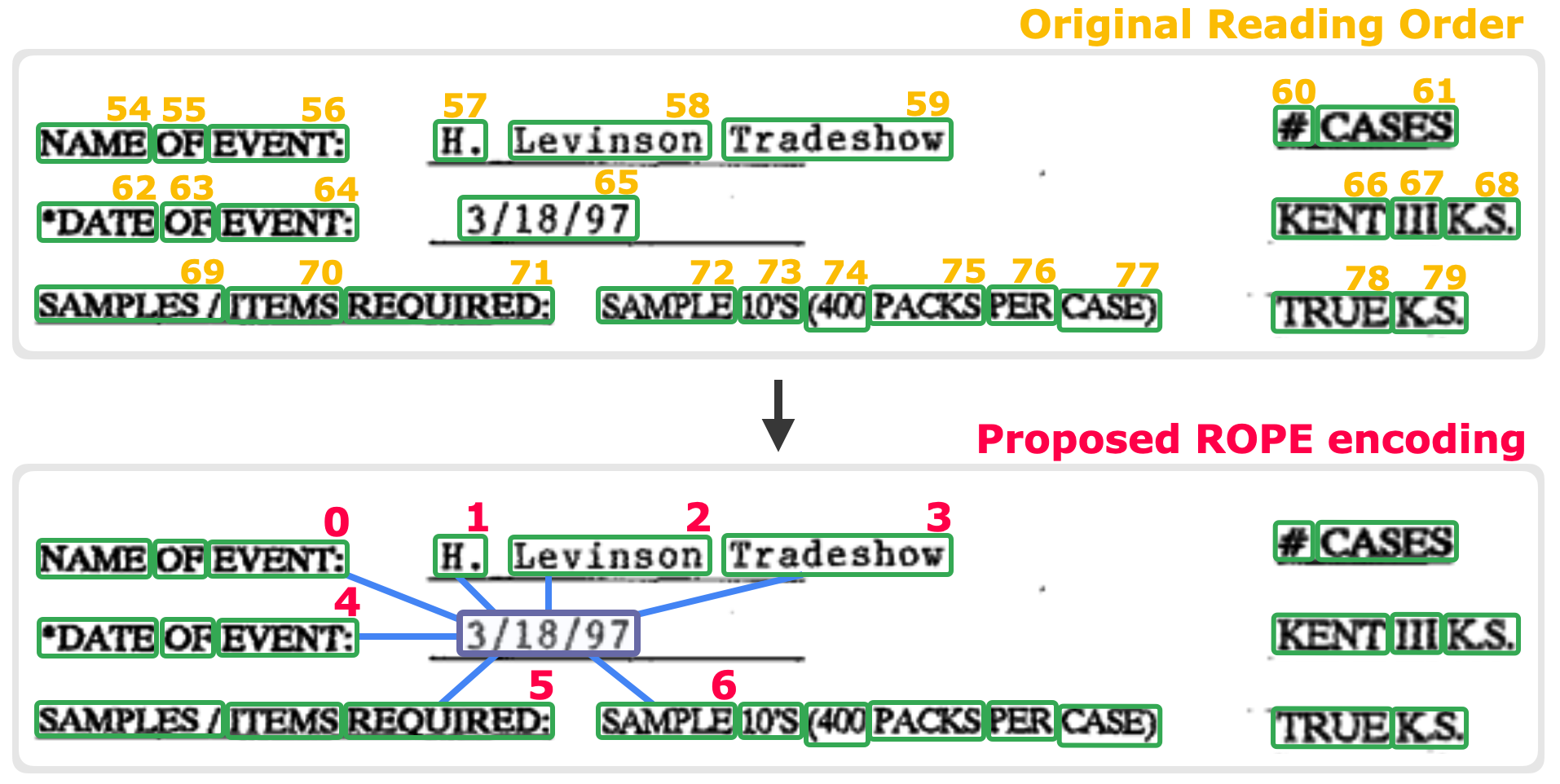}
\vspace{-2mm}
\caption{Illustration of the proposed Reading Order Equivariant Positional Encoding (ROPE). \textbf{Top:} a portion of a form document with the original word reading order. \textbf{Bottom:} given a graph connectivity, ROPE generates equivariant reading order codes with respect to the target word (in this case the date ``3/18/97'').} 
\label{fig:rope}
\end{figure}

Key information extraction from form-like documents
is one of the fundamental tasks of document understanding that has many real-world applications.
However, the major challenge of solving the task lies in modeling various template layouts and formats of documents. 
For example, a single document may contain multiple columns, tables, and non-aligned blocks of texts (e.g. Figure~\ref{fig:rope}).

The task has been studied from rule-based models~\cite{lebourgeois1992fast} to learning-based approaches~\cite{palm2017cloudscan,tata2021glean}. 
Inspired by the success of %named entity recognition in
sequence tagging in
NLP~\cite{sutskever2014sequence,vaswani2017attention,devlin2018bert}, 
a natural extension is applying these methods on linearly serialized 2D documents~\cite{palm2017cloudscan,aggarwal2020form2seq}.
Nevertheless, scattered columns, tables, and text blocks in documents make the serialization extremely difficult, largely limiting the performance of sequence models.
\citet{katti2018chargrid,zhao2019cutie} explore to directly work on 2D document space using grid-like convolutional models to better preserve spatial context during learning, but the performance is restrictive to the resolution of the grids.
Recently, \citet{qian2018graphie,davis2019deep,liu2019graph} propose to represent documents using graphs, where nodes define word tokens and edges describe the spatial patterns of words. \citet{yu2020pick} show state-of-the-art performance of Graph Convolutional Networks (GCNs)~\cite{duvenaud2015convolutional} on document understanding.

Although GCNs capture the relative spatial relationships between words through edges, the specific word ordering information is lost during the graph aggregation operation, in the similar way to the average pooling in Convolutional Neural Networks (CNNs).
However, we believe reading orders are strong prior to comprehending languages.
In this work, we propose a simple yet effective Reading Order Equivariant Positional Encoding (ROPE) that embeds the relative reading order context into graphs, bridging the gap between sequence and graph models for robust document understanding.
Specifically, for every word in a constructed graph, ROPE generates unique reading order codes for its neighboring words based on the graph connectivity. The codes are then fed into GCNs with self-attention aggregation functions for effective relative reading order encoding.
We study two fundamental entity extraction tasks including word labeling and word grouping on the public FUNSD dataset and a large-scale payment dataset. 
We observe that by explicitly encoding relative reading orders, ROPE brings the same or higher performance improvement compared to spatial relationship features in existing GCNs in parallel.

\section{Other Related Work}
Attention models show state-of-the-art results in graph learning~\cite{velivckovic2017graph} and NLP benchmarks~\cite{vaswani2017attention}. As attention models with positional encodings are proven to be universal approximators of sequence-to-sequence functions ~\cite{yun2019transformers}, encoding positions or ordering is an important research topic. For sequence, learned positional embeddings~\cite{gehring2017convolutional,devlin2018bert,shaw2018self}, sinusoidal functions and its extensions~\cite{liu2020learning} have been studied. Beyond that, positional encodings are explored in graphs~\cite{you2019position}, 2D images~\cite{parmar2018image} and 3D structures~\cite{fuchs2020se}. 
Lastly, graph modeling is also applied to other document understanding tasks, including document classification~\cite{yao2019graph} and summerization~\cite{yasunaga2017graph}.

\begin{figure}[!t]
\centering
\includegraphics[width=0.45\textwidth]{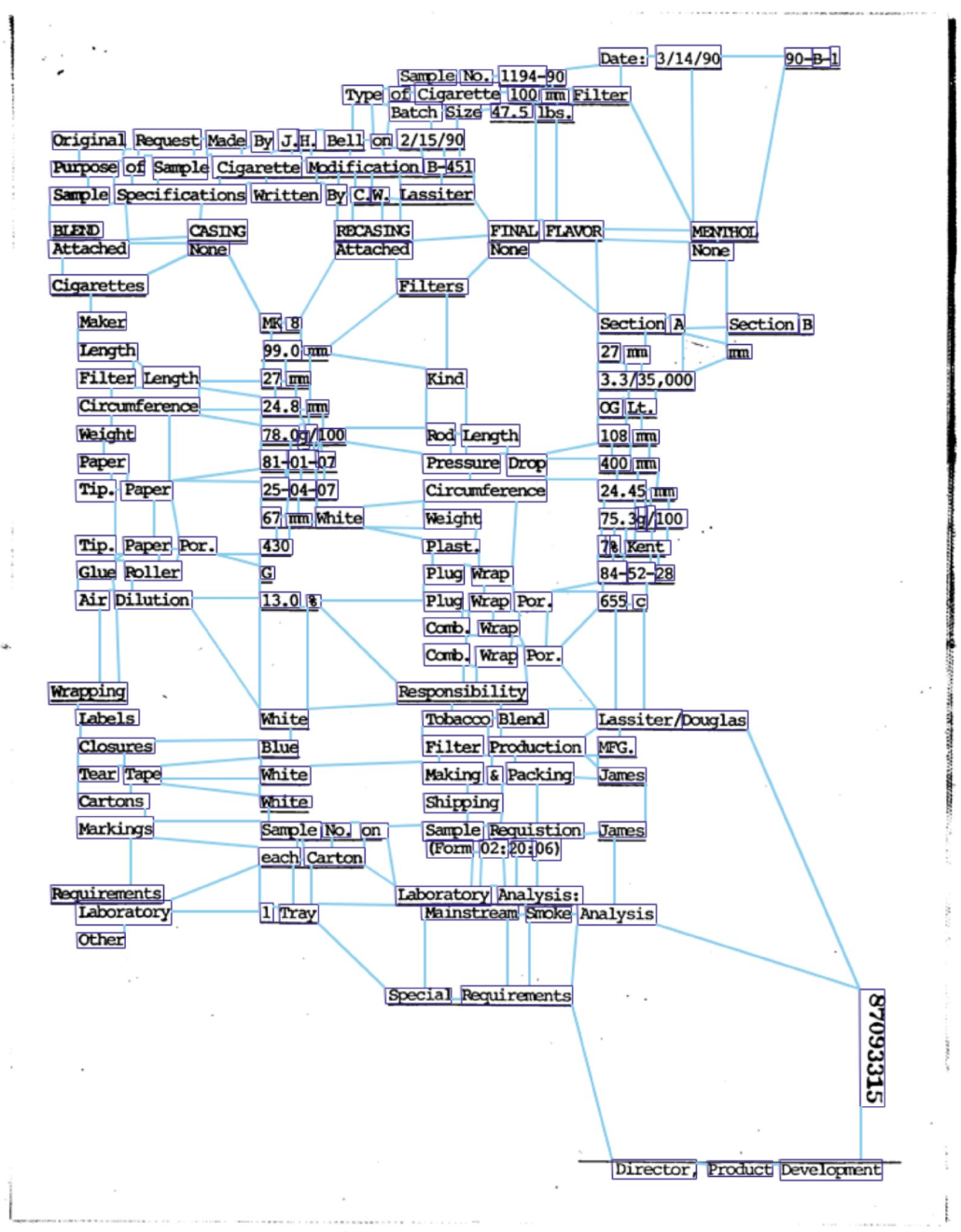}
\vspace{-2mm}
\caption{Sample of a $\beta$-skeleton graph of a document of FUNSD.}
\label{fig:graph_sample}
\end{figure}

\section{Method}
We follow recent advances in using GCNs for document information extraction that relax any serialization assumptions by sequence modeling.
GCNs take inputs (word tokens in this case) of arbitrary numbers, sizes, shapes and locations, and encode the underlying spatial layout patterns of documents through direct message passing and gradient updates between input embedding in the 2D space.

\begin{figure*}[!t]
\centering
\includegraphics[scale=0.67]{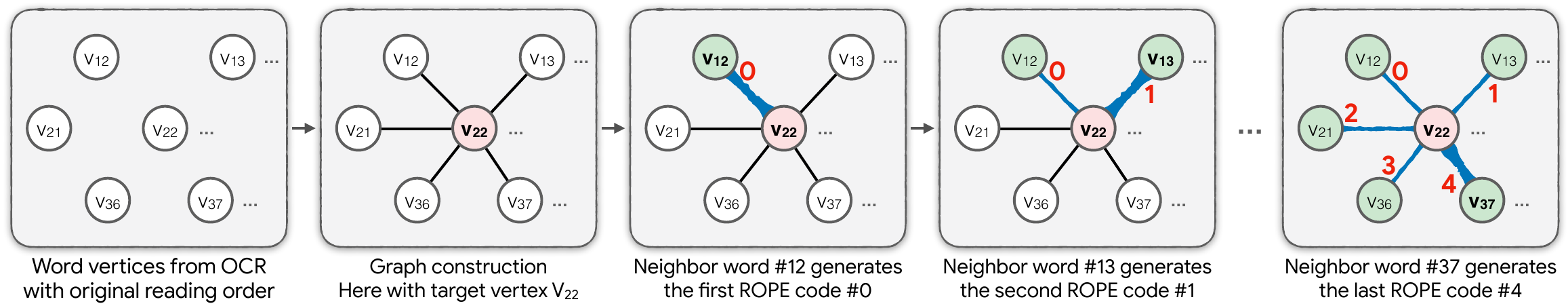}
\vspace{-2mm}
\caption{Implementation of the proposed Reading Order Equivariant Positional Encoding (ROPE). Given a graph connectivity, ROPE iterates through the neighboring word vertices in the original reading order and assigns new ROPE codes (red numbers) to the neighbors, starting from zero. Note that the proposed ROPE codes remain  unchanged if the neighbors and the target shift equally in the document with the same relative reading order, therefore being equivariant.}
\vspace{-2mm}
\label{fig:steps}
\end{figure*}

\paragraph{Node definition.}
Given a document $D$ with $N$ tokens denoted by $T = \{t_1, t_2, .. t_N\}$, we refer $t_i$ to the $i$-th token in a linearly serialized text sequence returned by the Optical Character Recognition (OCR) engine. The OCR engine generates the bounding box sizes and locations for all tokens, as well as the text within each box. We define node input representation for all tokens $T$ as vertices $V = \{v_1, v_2, .. v_N\}$, where $v_i$ concatenates quantifiable attributes available for  $t_i$. In our design, we use two common input modalities: (a) word embeddings from an off-the-shelf pre-trained BERT model~\cite{devlin2018bert}, and (b) spatial embeddings from normalized bounding box heights, widths, and Cartesian coordinate values of four corners.

\paragraph{Edge definition.}
While the vertices $V$ represent tokens in a document, the edges characterize the relationship between the vertices. Precisely, we define directional edges for a set of edges $E$, where each edge $e_{ij}$ connects two vertices $v_i$ and $v_j$, concatenating quantifiable edge attributes. In our design, we use two input modalities given an edge $e_{ij}$ connecting two vertices:
(a) spatial embeddings from horizontal and vertical normalized relative distances between centers, top left corners and bottom right corners of the bounding boxes. It also contains height and width aspect ratios of $v_i$, $v_j$, and relative height and width aspect ratios between $v_i$ and $v_j$.
(b) Visual embeddings that utilizes ImageNet pre-trained MobileNetV3~\cite{howard2019searching} to extract visual representations of union bounding boxes containing $v_i$ and $v_j$. 
The visual embedding in edge formation picks up visual cues such as colors, fonts, separating symbols or lines between two token bounding boxes (through their union bounding box).
We refer to the spatial embedding in (a) as the edge geometric (EdgeGeo) feature used in the experimental section.

\paragraph{Graph construction.}
Our implementation is based on the $\beta$-skeleton graph~\cite{kirkpatrick1985framework} with $\beta=1$ for graph construction. By using the ``ball-of-sight'' strategy, $\beta$-skeleton graph offers high connectivity between word vertices for necessary message passing while being much sparser than fully-connected graphs for efficient forward and backward computations~\cite{wang2021general}. A $\beta$-skeleton graph example can be found in Figure~\ref{fig:graph_sample}, and more can be found in   Figure~\ref{fig:beta_skeleton} in the Appendix.

\paragraph{Aggregation function.}
Inspired by the Graph Attention Networks~\cite{velivckovic2017graph} and the Transformers~\cite{vaswani2017attention}, we use multi-head self-attention module as our GCN aggregation (pooling) function. It calculates the importance of individual message coming from its neighbors to generate the new aggregated output.

\subsection{Reading Order Equivariant Positional Encoding (ROPE)}

Positional encoding~\cite{gehring2017convolutional} in 
sequence models is with an assumption that the input is perfectly serialized. However, as illustrated in Figure~\ref{fig:rope}, form-like documents often contain multiple columns or sections. A simple left-to-right and top-to-bottom serialization commonly provided by OCR engines does not provide accurate sequential presentation of words -- two consecutive words in the same sentence might have drastically different reading order indexes by naive serialization.

Instead of assigning absolute reading order indexes for the entire document at the beginning, we propose to encode the relative reading order context of neighboring words w.r.t. the target word based on the given graph connectivity. Figure~\ref{fig:steps} demonstrates the process of the proposed method: ROPE iterates through the neighboring word vertices in the original reading order and assigns new ROPE codes $p \in \NN$ (red numbers) to the neighbors, starting from zero. 
The generated codes are then appended to the corresponding incoming messages during graph message passing. Hence, ROPE provides a relative reading order context of the neighborhood for order-aware self-attention pooling.

Note that the generated ROPE codes remain unchanged if the neighbors and the target shift equally in the document with the same relative order, therefore being equivariant. Additionally, ROPE provides robust sequential output that is consistent even when the neighborhood crosses multiple columns or sections in a document.

Finally, we also explore sinusoidal encoding matrix~\cite{vaswani2017attention} besides the index-based encoding. Our ablation study in Section~\ref{sec:exp} shows that using both 
results in the best performance.

\section{Experiments}
\label{sec:exp}
We evaluate how reading order impacts overall performance of graph-based information extraction from form-like documents. We adopt  two form understanding tasks as~\citet{jaume2019}, including word labeling and word grouping. Word labeling is the task of assigning each word a label from a set of predefined entity categories, realized by node classification. Word grouping is the task of aggregating words that belong to the same entity, realized by edge classification. These two fundamental entity extraction tasks do not rely on perfect entity word groupings provided by the dataset and therefore help decouple the modeling capability provided by the proposed ROPE in practice.
These two tasks also effectively demonstrate the quality of the node embedding and edge embedding of the proposed graph architecture and decouple any performance gain from sophisticated Conditional Random Field (CRF) decoders often used on top of the model.

\subsection{Datasets}
\paragraph{Payment.}
We follow \citet{majumder2020representation} to prepare a large-scale payment document collection that consists of around 18K single-page payments. The data come from different vendors with different layout templates. For both word labeling and word grouping experiments, we use a 80-20 split of the corpus as the training and test sets.

We use a public OCR service\footnote{cloud.google.com/vision} to extract words from the payment documents. The service generates the text of each word with their corresponding 2D bounding box. The word boxes are roughly arranged in an order from left to right and from top to bottom. We then ask human annotators to label the words with 13 semantic entities.
Each entity ground truth is described by an entity type and a list of words generated by the OCR engine, resulting in over 3M word-level annotations. Labelers are instructed to label all instances of a field in a document, therefore our GCNs are trained to predict all instances of a field as well.

\paragraph{FUNSD.}
FUNSD~\cite{jaume2019} is a public dataset for form understanding in noisy scanned documents, containing a collection of research, marketing, and advertising documents that vary widely in their structure and appearance. The dataset consists of 199 annotated forms with 9,707 entities and 31,485 word-level annotations for 4 entity types: header, question, answer, and other. For both word labeling and word grouping experiments, we use the official 75-25 split for the training and test sets.

\subsection{Experimental Setup}
All GCN variants used in the experiment have the same architecture: The node update function is a 2-layer Multi-Layer Perceptron (MLP) with 128 hidden nodes. The aggregation function uses a 3-layer multi-head self-attention pooling with 4 heads and 32 as the head size.
The number of hops in the GCN is set to 7 for payment dataset and 2 for FUNSD dataset due to the complexity and scale of the former.
We use cross-entropy loss for both multi-class word labeling and binary word grouping tasks.
We train the models from scratch using Adam optimizer with the batch size of 1. The learning rate is set to 0.0001 with warm-up proportion of
0.01. The training is conducted on 8 Tesla P100 GPUs for approximately 1 day on the largest corpus.

\subsection{Results}
We train the GCNs from scratch on all datasets.
For word labeling we use multi-class node classification F1-scores as the metric and for word grouping we use binary edge classification F1-scores as the metric with the corresponding precision and recall values.

\begin{table}[!t]
\setlength{\tabcolsep}{7pt} % Default value: 6pt
\centering
\small
\resizebox{0.49\textwidth}{!}{
\begin{tabular}{lcccccc}
\toprule
 & \multicolumn{2}{c}{\textbf{Types of}} & \textbf{Word} & \multicolumn{3}{c}{\textbf{Word}} \\ 
                & \multicolumn{2}{c}{\textbf{Positional Encoding}} & \textbf{Labeling} & \multicolumn{3}{c}{\textbf{Grouping}} \\ 
&  & (ours) &   &   &  &     \\
 & EdgeGeo & ROPE & F1  & P & R & F1   \\
\midrule
\multirow{4}{*}{\rotatebox{90}{Payment}} &            &            & 60.80 & 83.64 & 83.97 & 83.80 \\
                    & \checkmark &            & 66.09 & 84.96 & 84.93 & 84.94 \\
                    &            & \checkmark & 68.17 & 84.92 & \textbf{86.86} & 85.88 \\
                    & \checkmark & \checkmark & \textbf{74.55} & \textbf{86.75} & 86.53 & \textbf{86.64} \\
\midrule
\multirow{4}{*}{\rotatebox{90}{FUNSD}} &            &            & 50.86 & 82.09 & 92.21 & 86.86 \\
                    & \checkmark &            & 53.16 & 87.56 & 87.17 & 87.37 \\
                    &            & \checkmark & 51.78 & \textbf{88.90}  & 89.67 & 89.28 \\
                    & \checkmark & \checkmark & \textbf{57.22} & 88.64 & \textbf{90.03} & \textbf{89.33} \\
\bottomrule
\end{tabular}
}
\vspace{-2mm}
\caption{\label{pos-enc-types} Different positional encodings for GCNs on information extraction tasks. We observe that the reading order encoding (ROPE) is equally or more important compared to edge geometric feature (EdgeGeo).}
\end{table}

\begin{table}[!t]
\setlength{\tabcolsep}{7pt} % Default value: 6pt
\centering
\small
\resizebox{0.49\textwidth}{!}{
\begin{tabular}{lcccccc}
\toprule
 & \multicolumn{2}{c}{\textbf{ROPE}} & \textbf{Word} & \multicolumn{3}{c}{\textbf{Word}} \\ 
                & \multicolumn{2}{c}{\textbf{Encoding Function}} & \textbf{Labeling} & \multicolumn{3}{c}{\textbf{Grouping}} \\ 
 & Index & Sinusoidal & F1  & P & R & F1   \\
\midrule
\multirow{4}{*}{\rotatebox{90}{Payment}} &  &   & 66.09 & 84.96 & 84.93 & 84.94 \\
                    & \checkmark &              & 72.41 & \textbf{87.78} & 85.31 & 86.53 \\
                    &            & \checkmark   & 70.94 & 88.49 & 83.00 & 85.66 \\
                    & \checkmark & \checkmark   & \textbf{74.55} & 86.75 & \textbf{86.53} & \textbf{86.64} \\
\midrule
\multirow{4}{*}{\rotatebox{90}{FUNSD}} &  &   & 53.16 & 87.56 & 87.17 & 87.37 \\
                    & \checkmark &            & 55.48 & 85.95 & \textbf{92.15} & 88.94 \\
                    &            & \checkmark & 54.14 & \textbf{88.72} & 89.51 & 89.12 \\
                    & \checkmark & \checkmark & \textbf{57.22} & 88.64 & 90.03 & \textbf{89.33} \\
\bottomrule
\end{tabular}
}
\vspace{-2mm}
\caption{\label{enc-func} Ablation of positional encoding function used in the proposed ROPE. We observe that either index or sine encoding works better than no positional encoding. Combined works the best.}
\end{table}

\paragraph{Importance of reading order.}
Positional encoding mechanisms are the key components to exploiting layout patterns of words -- {\tt Answer} entities are usually next to or below the {\tt Question} entities. Existing GCN approaches rely on edge geometric (EdgeGeo) features to capture such spatial relationships between words in 2D space. Here we evaluate the importance of the proposed reading order encoding ROPE with various combinations of EdgeGeo over the baseline GCN~\cite{qian2018graphie} as summarized in Table~\ref{pos-enc-types}. Without any positional encoding, word labeling F1 drops by 13.75 points and word grouping F1 drops by 2.84 points on payment dataset.  Then, we pass ROPE to incoming messages and find that this reduces the drop to 6.38 points on word labeling and 0.76 points on word grouping. Similar trend can be observed on FUNSD as well. Surprisingly, ROPE reduces performance drop more effectively than EdgeGeo on the larger payment dataset. Given these ablations, we conclude that reading order information is at least the same or more important than geometric features, and they bring orthogonal improvements to the overall performance.

\paragraph{Reading order encoding function.}
In practice, each target word usually has less than 8 neighboring words given a constructed $\beta$-skeleton graph. Therefore, a natural approach to assigning relative reading orders is to simply use the ROPE encoded indexes. In Table~\ref{enc-func} we observe that simple index encoding immediately improves GCN without ROPE by 6.32 points on word labeling and 1.59 points on word grouping using payment corpus. Next we explore the popular sinusoidal function (with 3 base frequencies) for reading order encoding. It improves GCN without ROPE by 4.85 points on word labeling and 0.72 points on word grouping. Interestingly, sine function provides on par performance but does not outperform index encoding. The reason might be because the $\beta$-skeleton graph does not generate an extremely large number of neighbors, so simple index encoding is sufficient.

\begin{figure}[!t]
\centering
\includegraphics[scale=0.3]{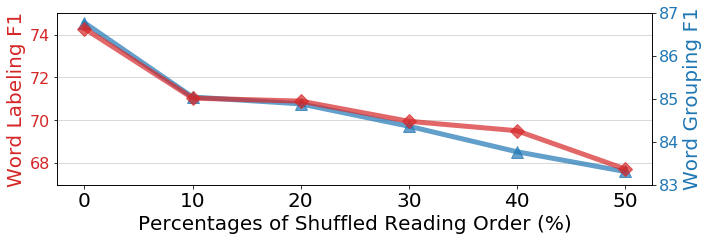}
\vspace{-2mm}
\caption{Sensitivity of ROPE to OCR reading order on Payment. The proposed ROPE codes remain the same if the connected neighboring words and target word shift equally in the document.}
\label{fig:shuffle}
\end{figure}

\paragraph{Sensitivity to OCR reading order.}
We investigate the robustness of ROPE to the quality of the input reading order.
We shuffle the reading order provided by the OCR engine with a varying percentage of words before feeding into ROPE.
Figure~\ref{fig:shuffle} exhibits the performance. For both word labeling and word grouping tasks, ROPE provides performance improvement up to less than 30\% word order shuffling on the large payment corpus. With 30\% or more word order shuffled, we observe less performance degradation on the word labeling, suggesting that the word grouping task is more sensitive to the original OCR reading order.

\section{Conclusion}
We present a simple and intuitive reading order encoding method ROPE that is equivariant to relative reading order shifting.
It embeds the effective positional encoding from sequence models while leveraging the existing spatial layout modeling capability of graphs. 
We foresee the proposed ROPE can be immediately applicable to other document understanding tasks.

\paragraph{Acknowledgements.}
We are grateful to Evan Huang, Lauro Beltrão Costa, Yang Xu, Sandeep Tata, and Navneet Potti for the helpful feedback on this work.

\bibliographystyle{acl_natbib}
\bibliography{egbib}

\newpage
\clearpage

\appendix

\begin{figure*}[!ht]
\centering
\includegraphics[width=1\textwidth]{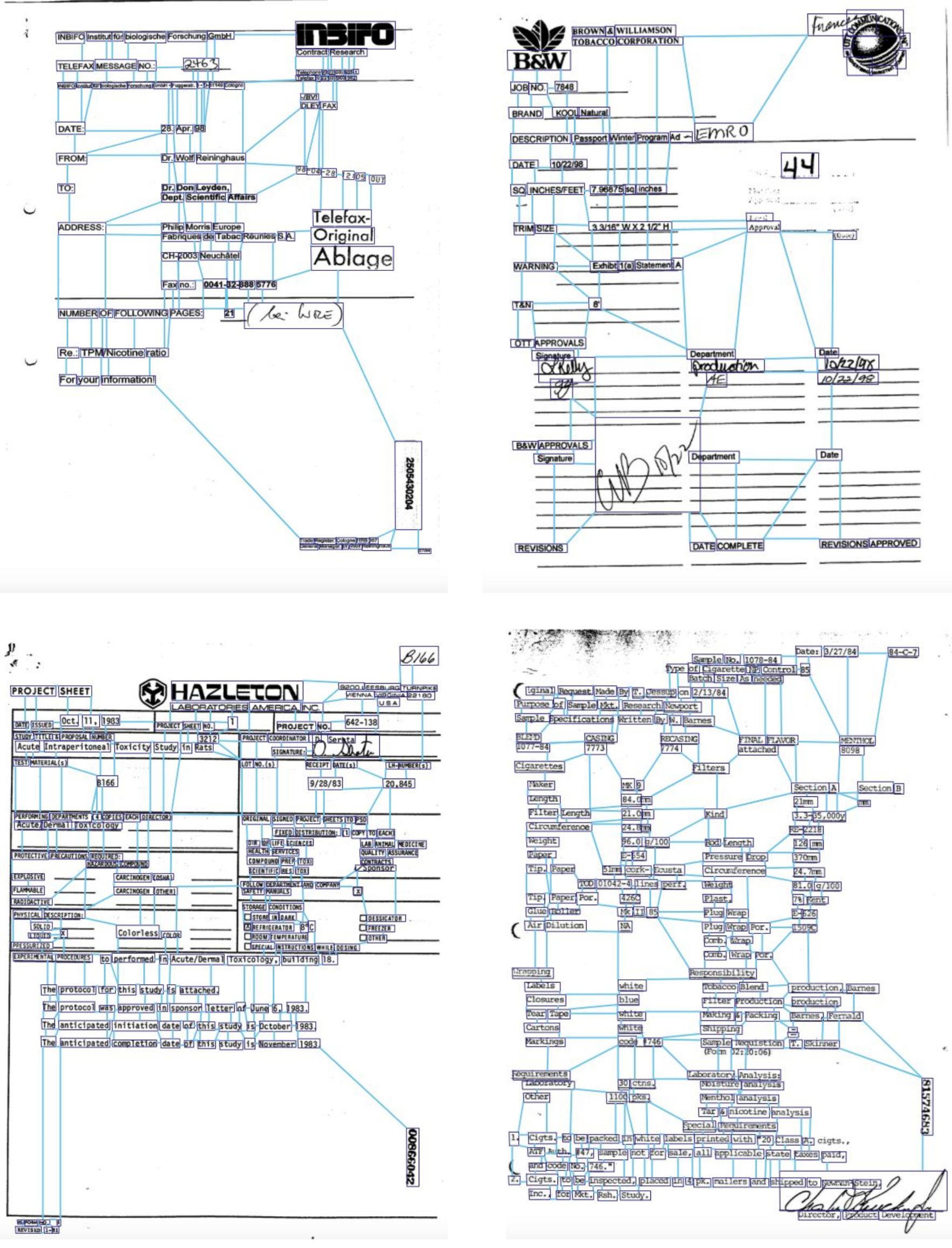}
\caption{$\beta$-skeleton examples of documents of FUNSD. By using the ``ball-of-sight'' strategy, $\beta$-skeleton graph offers high connectivity between word vertices for necessary message passing while being much sparser than fully-connected graphs for efficient forward and backward computations}
\label{fig:beta_skeleton}
\end{figure*}

\begin{figure*}[!ht]
\centering
\includegraphics[width=1\textwidth]{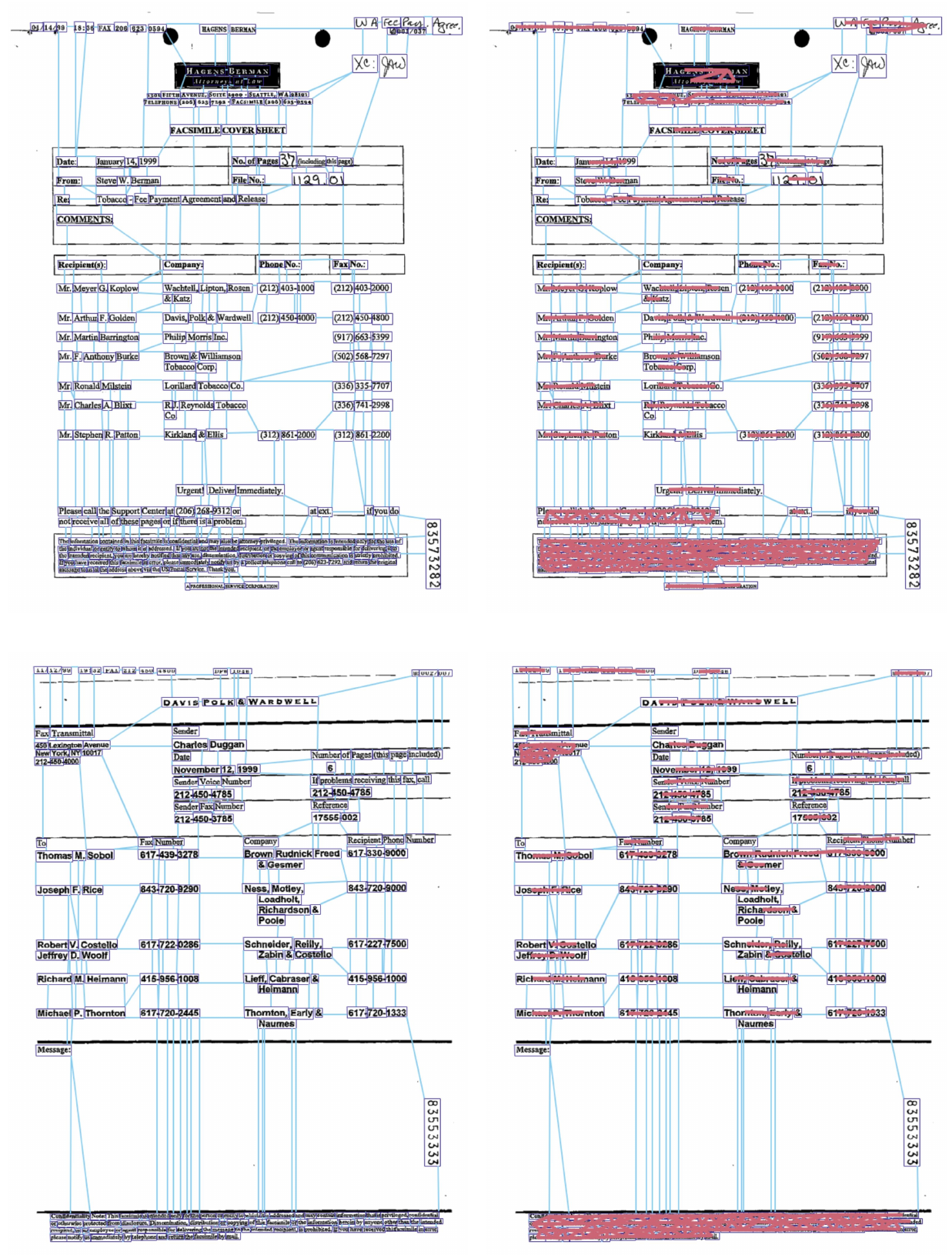}
\caption{Sample output of the word grouping task on FUNSD with a few failure cases.}
\label{fig:failure}
\end{figure*}

\end{document}